\documentclass[10pt,twocolumn,letterpaper]{article}
\usepackage{iccv}
\pdfoutput=1
\usepackage{times}
\usepackage{epsfig}
\usepackage{graphicx}
\usepackage{amsmath}
\usepackage{amssymb}
\usepackage{comment}
\usepackage{xspace}
\usepackage{color}
\usepackage{multirow}
\usepackage{pdfpages}
\usepackage[numbers,sort&compress]{natbib}
\makeatletter
\DeclareRobustCommand\onedot{\futurelet\@let@token\@onedot}
\def\@onedot{\ifx\@let@token.\else.\null\fi\xspace}
\def\eg{\emph{e.g}\onedot} 

\def\ie{\emph{i.e}\onedot}

\def\etc{\emph{etc}\onedot}

\def\etal{\emph{et al}\onedot}
\makeatother

\usepackage[breaklinks=true,bookmarks=false]{hyperref}

\iccvfinalcopy 

\def\iccvPaperID{****} 

\ificcvfinal\fi

\begin{document}

\title{Aesthetic Image Captioning From Weakly-Labelled Photographs}

\author{Koustav Ghosal\\
\and
Aakanksha Rana\\
\and
Aljosa Smolic\\
\and
V-SENSE, School of Computer Science and Statistics, Trinity College Dublin, Ireland.\\
}

\maketitle
\ificcvfinal\thispagestyle{empty}\fi

\begin{abstract}
Aesthetic image captioning (AIC) refers to the multi-modal task of generating
critical textual feedbacks for photographs.  
While in natural image captioning (NIC), deep models are trained in an end-to-end manner using large curated datasets such as MS-COCO, no such large-scale, clean dataset exists for AIC. Towards this goal, we propose an automatic cleaning strategy to create a benchmarking AIC dataset, by exploiting the images and noisy comments easily available from photography websites. We propose a probabilistic caption-filtering method for cleaning the noisy web-data, and compile a large-scale, clean dataset `AVA-Captions', ( $\sim230,000$ images with $\sim5$ captions per image). Additionally, by exploiting the latent associations between aesthetic attributes, we propose a strategy for training a convolutional neural network (CNN) based visual feature extractor, typically the first component of an AIC framework. The strategy is weakly supervised and can be effectively used to learn rich aesthetic representations, without requiring expensive ground-truth annotations. 
We finally showcase a thorough analysis of the proposed contributions using automatic metrics and subjective evaluations.
\end{abstract}

\section{Introduction}
\label{sec:introduction}

\begin{figure*}
\begin{center}
\resizebox{\textwidth}{!}{
\def\arraystretch{1.75}
\begin{tabular}{p{.175\textwidth}|p{.25\textwidth}p{.25\textwidth}p{.25\textwidth}p{.225\textwidth}}
\multicolumn{1}{p{.175\textwidth}}{\textbf{Training Strategy}} & \includegraphics[width=0.2\textwidth, height=0.15\textwidth ]{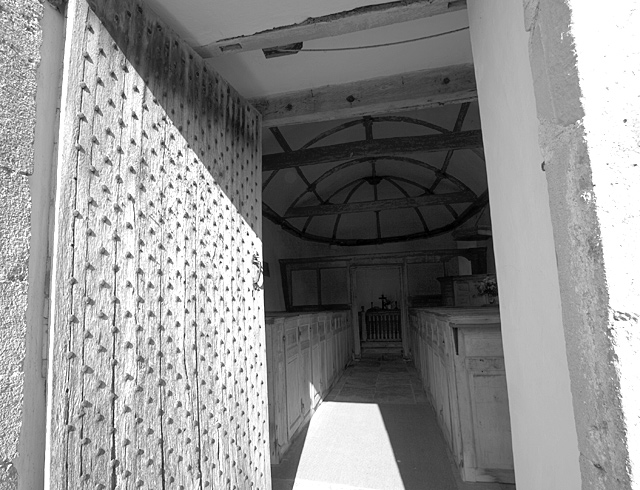} & \includegraphics[width=0.2\textwidth, height=0.15\textwidth]{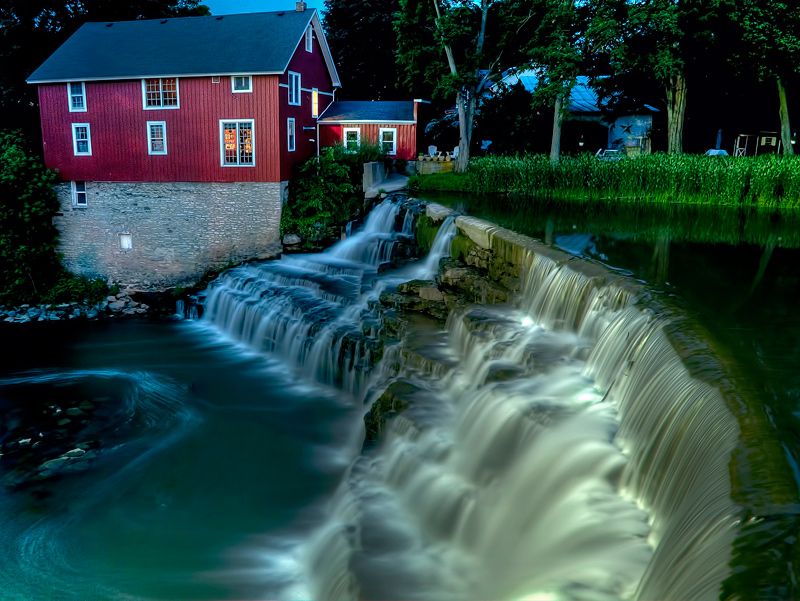} & \includegraphics[width=0.2\textwidth, height=0.15\textwidth]{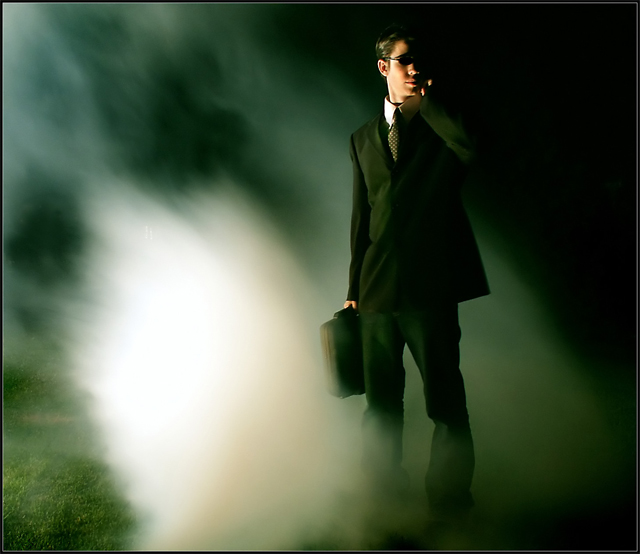} & \includegraphics[width=0.2\textwidth, height=0.15\textwidth]{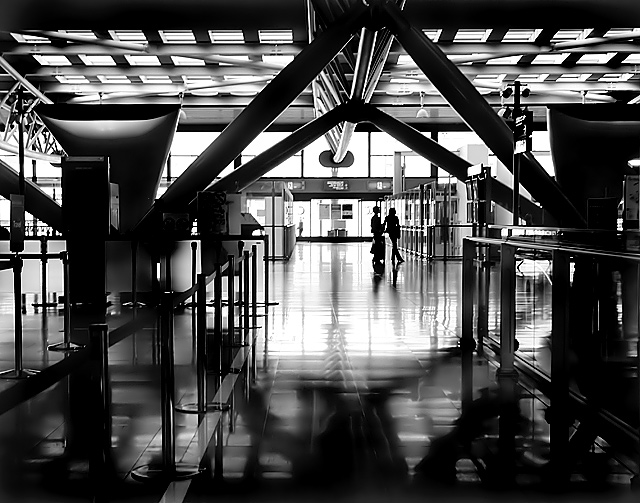} \tabularnewline
\hline 
\textbf{(a)} Noisy Data \& Supervised CNN \textbf{(NS)} & i like the angle and the composition & i like the colors and the composition & i like the composition and the lighting & i like the composition and the bw \tabularnewline
\hline 
\textbf{(b)} Clean Data \& Supervised CNN \textbf{(CS)} & i like the idea , but i think it would have been better if the door
was in focus . & i like the colors and the water . the water is a little distracting
. & i like the way the light hits the face and the background . & i like this shot . i like the way the lines lead the eye into the
photo .\tabularnewline
\hline 
\textbf{(c)} Clean Data \& Weakly Supervised CNN \textbf{(CWS)} & i like the composition , but i think it would have been better if
you could have gotten a little more of the building & i like the composition and the colors . the water is a little too
bright . & this is a great shot . i love the way the light is coming from the
left . & i like the composition and the bw conversion . \tabularnewline
\hline 
\end{tabular}}
\end{center}
\caption{Aesthetic image captions. We show candidates generated by three different frameworks discussed in this paper: \textbf{(a)} For NS, we use an ImageNet trained CNN  and LSTM  trained on noisy comments \textbf{(b)} For CS, we use an ImageNet trained CNN  and LSTM  trained on compiled AVA-Captions dataset \textbf{(c)} For CWS, we use a weakly-supervised CNN and LSTM trained on AVA-Captions}
\label{fig:title_fig}
\end{figure*}

Availability of large curated datasets such as MS-COCO~\cite{lin2014microsoft} ($100K$ images), Flickr30K~\cite{plummer2015flickr30k} ($30K$ images) or Conceptual Captions \cite{sharma2018conceptual} ($3M$ images) made it possible to train deep learning models for complex, multi-modal tasks such as natural image captioning (NIC)~\cite{xushow} where the goal is to factually describe the image content. Similarly, several other captioning variants such as visual question answering \cite{antol2015vqa}, visual storytelling \cite{Kiros2015}, stylized captioning \cite{mathews2018semstyle} have also been explored. Recently, the PCCD dataset ($\sim4200$ images) ~\cite{chang2017aesthetic} opened up a new area of research of describing images aesthetically. Aesthetic image captioning (AIC) has potential applications in the creative industries such as developing smarter cameras or web-based applications, ranking, retrieval of images and videos \etc. However in~\cite{chang2017aesthetic}, only six well-known photographic/aesthetic attributes such as composition, color, lighting, \etc have been used to generate aesthetic captions with a small curated dataset. Hence, curating a large-scale dataset to facilitate a more comprehensive and generalized understanding of aesthetic attributes remains an open problem.

Large-scale datasets have always been pivotal for research advancements in various fields~\cite{deng2009imagenet,lin2014microsoft,plummer2015flickr30k,ra2019TIP}. However, manually curating such a dataset for AIC is not only time consuming, but also difficult due to its subjective nature. Moreover, a lack of unanimously agreed `standard' aesthetic attributes makes this problem even more challenging as compared to its NIC counterpart, where deep models are trained with known attributes/labels~\cite{lin2014microsoft}. In this paper, we make two contributions. Firstly, we propose an automatic cleaning strategy to generate a large scale dataset by utilizing the noisy comments or aesthetic feedback provided by users for images on the web.
Secondly, for a CNN-based visual feature extractor as is typical in NIC pipelines, we propose a weakly-supervised training strategy. By automatically discovering certain `meaningful and complex aesthetic concepts', beyond the classical concepts such as composition, color, lighting, \etc, our strategy can be adopted in scenarios where finding clean ground-truth annotations is difficult (as in the case of many commercial applications). We elaborate these contributions in the rest of this section. 

To generate a clean aesthetic captioning dataset, we collected the raw user comments from the Aesthetic Visual Analysis (AVA) dataset \cite{murray2012ava}. AVA is a widely used dataset for aesthetic image analysis tasks such as aesthetic rating prediction \cite{lu2014rapid, Ma_2017_CVPR}, photographic style classification \cite{Karayev2014, ghosalgeometry}. However, AVA was not created  for AIC. In this paper, we refer to the original AVA with raw user comments as AVA raw-caption. It contains $\sim250,000$ photographs from dpchallenge.com and the corresponding user comments or feedback for each photograph ($~3$ billion in total). Typically, in Dpchallenge, users ranging from casual hobbyists to expert photographers provide feedback to the images submitted and describe the factors that make a photograph aesthetically pleasing or dull. Even though these captions contain crucial aesthetic-based information from images, they cannot be directly used for the task of AIC. Unlike the well instructed and curated datasets~\cite{lin2014microsoft}, AVA raw-captions are unconstrained user-comments in the wild with typos, grammatically inconsistent statements, and also containing a large number of comments occurring frequently without useful information. Previous work in AIC~\cite{chang2017aesthetic} acknowledges the difficulty of dealing with the highly noisy captions available in AVA.

In this work, we propose to clean the raw captions from AVA by proposing a probabilistic n-gram based filtering strategy. Based on word-composition and frequency of occurrence of n-grams, we propose to assign an informativeness score to each comment, where comments with a little or vague information are discarded. Our resulting clean dataset, \textbf{AVA-Captions} contains $\sim230,000$ images and $\sim1.5M$ captions with an average of $\sim5$ comments per image and can be used to train the Long and Short Term Memory (LSTM) network in the image captioning pipeline in the traditional way. Our subjective study verifies that the proposed automatic strategy is consistent with human judgement regarding the informativeness of a caption. Our quantitative experiments and subjective studies also suggest that models trained on AVA-Captions are more diverse and accurate than those trained on the original noisy  AVA-Comments. It is important to note that our strategy to choose the large-scale AVA raw-caption is motivated from the widely used image analysis benchmarking dataset, MS-COCO, which is now used as an unified benchmark for diverse tasks such as object detection, segmentation, captioning, \etc. We hope that our cleaned dataset will serve as a new benchmarking dataset for various creative studies and aesthetics-based applications such as aesthetics based image enhancement, smarter photography cameras, \etc. 

Our second contribution in this work is a weakly supervised approach for training a CNN, as an alternative to the standard practice. The standard approach for most image captioning pipelines is to train a CNN on large annotated datasets \eg ImageNet \cite{deng2009imagenet}, where rich and discriminative visual features are extracted corresponding to the physical properties of objects such as cars, dogs etc. These features are provided as input to an LSTM for generating captions. Although trained for classification, these ImageNet-based features have been shown to translate well to other tasks such as segmentation \cite{long2015fully}, style-transfer \cite{gatys2016image}, NIC. In fact, due to the unavailability of large-scale, task-specific CNN annotations, these ImageNet features have been used for other variants of NIC such as aesthetic captioning \cite{chang2017aesthetic}, stylized captioning \cite{mathews2018semstyle}, product descriptions \cite{yashima2016learning}, \etc.

However, for many commercial/practical applications, availability of such datasets or models is unclear due to copyright restrictions \cite{meduim_annotation, quartz_annotation,racolor}. On the other hand, collecting task-specific manual annotations for a CNN is expensive and time intensive. Thus the question remains open if we can achieve better or at least comparable performance by utilizing easily available weak annotations from the web (as found in AVA) and use them for training the visual feature extractor in AIC.
To this end, motivated from weakly supervised learning methods \cite{doersch2017multi, ren2018cross}, we propose a strategy which exploits the large pool of unstructured raw-comments from AVA and discovers latent structures corresponding to meaningful \textit{photographic concepts} using Latent Dirichlet Allocation (LDA) \cite{blei2003latent}.
We experimentally observe that the weakly-supervised approach is effective and its performance is comparable to the standard ImageNet trained supervised features.

In essence, our contributions are as follows:
\begin{enumerate}
    \item We propose a caption filtering strategy and compile AVA-Captions, a large-scale and clean dataset for aesthetic image captioning (Sec \ref{subsec:Caption-filtering-strategy}). 
    \item We propose a weakly-supervised approach for training the CNN of a standard CNN-LSTM framework (Sec \ref{subsec:Discovering-Aesthetic-Attributes})
    \item We showcase the analysis of the AIC pipeline based on the standard automated metrics (such as BLEU, CIDEr,  SPICE \etc \cite{papineni2002bleu,vedantam2015cider,anderson2016spice}), diversity
of captions and subjective evaluations which are publicly available for further explorations (Section \ref{sec:Experiments}).
\end{enumerate}
\section{Related Work}
\label{sec:reltd}
Due to the multi-modal nature of the task, the problem spans into many different areas  of image and text analysis and thus related literature abound. However, based on the primary theme we roughly divide this section into four areas as follows: 

\textbf{Natural Image Captioning:} While early captioning methods ~\cite{hodosh2013framing,socher2014grounded,farhadi2010every,ordonez2011im2text,jia2011learning}
followed a dictionary look-up approach, recent parametric methods
~\cite{johnson2016densecap,fang2015captions,karpathy2015deep,mao2016generation,anne2016deep,mao2014deep,donahue2015long,mao2015learning,geman2015visual,malinowski2015ask,tu2014joint,bigham2010vizwiz,agrawal2018don,jin2015aligning}
are generative in the sense that they learn a mapping from visual to textual
modality. Typically in these frameworks, a CNN is followed by a RNN or LSTM \cite{xushow,johnson2016densecap,fang2015captions,karpathy2015deep,mao2016generation,anne2016deep,mao2014deep,donahue2015long,mao2015learning}, although fully convolutional systems have been proposed by Aneja \etal \cite{aneja2018convolutional} recently. 

\textbf{Image Aesthetics:} 
 Research in understanding the perceptual and aesthetic concepts in images can be divided
into the model-based ~\cite{datta2006studying,ke2006design,luo2008photo,obrador2012towards,dhar2011high,joshi2011aesthetics,san2012leveraging,aydin2015automated}
and the data-driven ~\cite{lu2015deep,lu2014rapid,Ma_2017_CVPR,mai2016composition,Karayev2014,1707.03981}
approaches. While model-based approaches rely on manually hard-coding
the aspects such as the Rule of Thirds, depth of field,
colour harmony, etc., the
data driven approaches usually train CNNs
on large-scale datasets and either predict
an overall aesthetic rating \cite{lu2014rapid,lu2015deep,mai2016composition}
or a distribution over photographic attributes ~\cite{Karayev2014,lu2014rapid,lu2015deep,ghosalgeometry}.

\textbf{Learning from Weakly-Annotated / Noisy Data:} Data dependency of very deep neural nets and the high cost of human supervision has led to a natural interest towards exploring the easily available web-based big data. Typically in these approaches, web-crawlers collect easily available noisy multi-modal data \cite{berg2010automatic, chen2013neil, vittayakorn2016automatic} or e-books \cite{divvala2014learning} which is jointly processed for labelling and knowledge extraction. The features are used for diverse applications such as classification and retrieval \cite{sun2015automatic,2014feature} or product description generation \cite{yashima2016learning}.

\textbf{Aesthetic Image Captioning:} To the best of our knowledge, the problem of aesthetic image captioning
has been first and only addressed by Chang \etal in ~\cite{chang2017aesthetic}.
The authors propose a framework which extracts features covering seven
different aspects such as general impression, composition and perspective,
color and lighting, etc. and generate meaningful captions by fusing
them together. They compile the photo critique captioning dataset (PCCD)
with $\sim4K$ images and $\sim30K$ captions. While their method is purely supervised and the network is trained using strong labels, we adopt a weakly-supervised approach to train our network with indirect labels. Additionally, AVA-Captions is a significantly bigger ($\sim60$ times) dataset with $\sim240K$ and $\sim1.3M$ images and captions, respectively. The scale of AVA allows training deeper and more complex architectures which can be generalized to PCCD as well. We demonstrate this later in Table \ref{tab:automated_metrics}b.
\section{Caption Filtering Strategy}
\label{subsec:Caption-filtering-strategy}
\begin{figure}
\begin{center}
\resizebox{.45\textwidth}{!}{
\def\arraystretch{1.2}
\begin{tabular}{p{0.15\textwidth}p{0.25\textwidth}|p{0.05\textwidth}}
 \hline
 \multicolumn{1}{c}{\textbf{Image}} & \multicolumn{1}{c}{\textbf{Comments}} & \multicolumn{1}{c}{\textbf{Scores}}\tabularnewline
\hline 
\hline
\multirow{7}{*}{\includegraphics[width=0.15\textwidth]{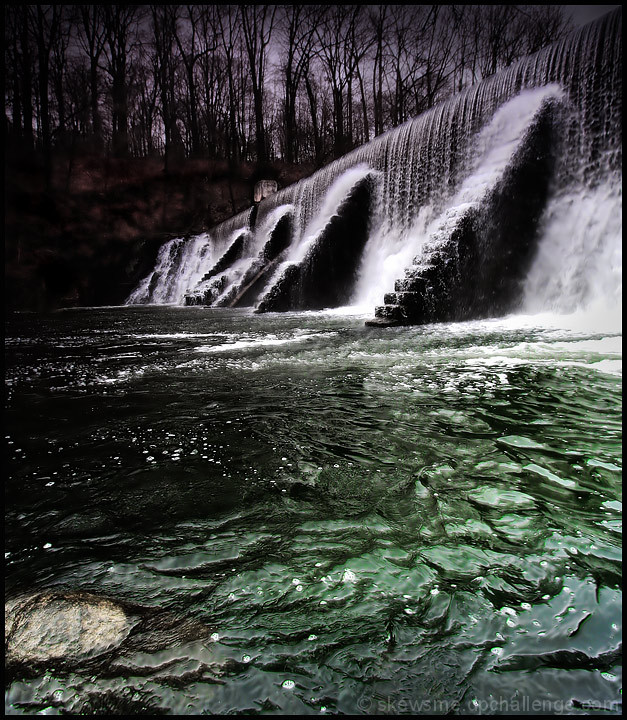}} &  Photo Quality : Awesome & $9.62$ \tabularnewline
\cline{2-3} 
 & I love the \textbf{colors} here  & $1.85$ \tabularnewline
\cline{2-3} 
 & I like the \textbf{trees} in the \textbf{background} and the \textbf{softness} of the \textbf{water}. & $28.41$ \tabularnewline
\cline{2-3} 
& The \textbf{post processing} \textbf{looks great} with the \textbf{water}, but the \textbf{top half} of the photo doesn't work as well. & $47.44$\tabularnewline
\hline 
\end{tabular}
}
\end{center}
\caption{Informativeness of captions. We suggest the readers to check the supplementary material for more comments and the corresponding scores.}
\label{fig:Informativeness-of-captions}
\end{figure}
 
In AVA raw-caption, we observe two main types of undesirable captions. First, there are
captions which suffer from generic noise frequently observed in most text corpora, especially those compiled from social media. They include typing errors, non-English comments, colloquial acronyms, exclamatory words (such as ``woooow''), extra punctuation (such as ``!!!!''), etc. Such noise can be handled using standard natural language processing techniques \cite{Loper:2002:NNL:1118108.1118117}. 

Second, we refer to the \textit{safe} comments, which carry a little or no useful
information about the photograph. For example, in Figure \ref{fig:Informativeness-of-captions}, comments such as ``\emph{Photo Quality : Awesome}'' or ``\emph{I love the colors here}'' provide a
valid but less informative description of the photograph . 
It is important to filter these comments, otherwise the network ends up learning these less-informative, \emph{safe} captions by ignoring the more informative and discriminative ones such as ``\emph{The post processing looks great with the water, but the top half of the photo doesn't work as well.}'' \cite{chang2017aesthetic}. 

To this end, we propose a probabilistic strategy for caption filtering based on the informativeness of a caption. Informativeness is measured by the presence of certain n-grams. The approach draws motivation from two techniques frequently used in vision-language problems --- word composition and term-frequency - inverse document frequency (TF-IDF).

\textbf{Word Composition:} Bigrams of the ``descriptor-object'' form often convey more information than the unigrams of the  objects alone. For example, ``post processing'' or ``top half'' convey more information than ``processing'' or ``half''. 
On the other hand, the descriptors alone may not always be sufficient to describe a complete concept and its meaning is often closely tied to the object \cite{nagarajan2018attributes}. For example, ``sharp`` could be used in two entirely different contexts such as ``sharp contrast'' and ``sharp eyes''. This pattern is also observed in the 200 bigrams (or ugly and beautiful attributes) discovered from AVA by Marchesotti \etal \cite{murray2012ava} such as ``nice colors'', ``beautiful scene'', ``too small'', ``distracting background'', \etc. Similar n-gram modelling is found in natural language processing as adjective-noun \cite{socher2014grounded, misra2017red, santa2018neural} or verb-object \cite{sadeghi2011recognition,zhang2017visual} compositions. 

\textbf{TF-IDF:}  The other motivation is based on the intuition
that the key information in a comment is stored in certain n-grams which occur less frequently in the comment corpus
such as ``softness'', ``post processing'', ``top half'' \etc. A sentence composed of frequently occurring n-grams such as ``colors" or ``awesome" is less likely to contain useful information. The intuition follows from the motivation of commonly used TF-IDF metric in document classification, which states that more frequent words of a vocabulary are less discriminative for document classification \cite{ramos2003using}. Such hypothesis also forms a basis in the CIDEr evaluation metric \cite{vedantam2015cider} widely used for tasks such as image captioning, machine translation, \etc.

\textbf{Proposed ``Informativeness'' Score:} Based on these two criteria, we start by constructing two vocabularies as follows: for unigrams we choose only the nouns and for bigrams we select ``descriptor-object'' patterns \ie where the first term is a noun, adjective or adverb and the second term is a noun or an adjective. Each n-gram $\omega$ is assigned a corpus probability $P$ as: 
\begin{equation}
P(\omega)=\frac{C_{\omega}}{\sum_{i=1}^{D}C_{i}}\label{eq:probability_tf}
\end{equation} 
where the denominator sums the frequency of each n-gram $\omega$ such that $\sum_{i=1}^{D} P(\omega_i) = 1$, where $D$ is the vocabulary size, and $C_\omega$ is the corpus frequency of n-gram $\omega$. Corpus frequency of an n-gram refers to the number of times it occurs in the comments from all the images combined. This formulation assigns high probabilities for more frequent words in the comment corpus. 

Then, we represent a comment as the union of its unigrams
and bigrams i.e., $S=(S_{u}\cup S_{b})$ , where $S_{u}=(u_{1}u_{2}\dots u_{N})$
and $S_{b}=(b_{1}b_{2}\dots b_{M})$ are the sequences of unigrams
and bigrams, respectively. A comment is assigned an informativeness
score $\rho$ as follows: 

\begin{equation}
\rho_{s}=-\frac{1}{2}[\log\prod_{i}^{N}P(u_{i})+\log\prod_{j}^{M}P(b_{j})]\label{eq:informativeness}
\end{equation}

where $P(u)$ and $P(b)$ are the probabilities of a unigram or bigram given by Equation \ref{eq:probability_tf}. Equation \ref{fig:Informativeness-of-captions}
is the average of the negative log probabilities of $S_{u}$ and $S_{b}$. 

Essentially, the score of a comment is modelled as the joint probability of n-grams in it, following the simplest Markov assumption \ie all n-grams are independent \cite{jurafsky2000speech}. If the n-grams in a sentence have higher corpus probabilities then the corresponding score $\rho$ is low due to the negative logarithm, and vice-versa.

Note that the score is the negative logarithm of the product of probabilities and longer captions tend to receive higher scores. However, our approach does not \textit{always} favour long comments, but does so only if they consist of ``meanigful'' n-grams conforming to the ``descriptor-object'' composition. In other words, randomly long sentences without useful information are discarded. On the other hand, long and informative comments are kept. This is also desirable as longer comments in AVA tend
to be richer in information as expert users are specifically asked
to provide detailed assessment which is referred to as \emph{critique club effect} in \cite{marchesotti2015discovering}.

We label a comment as informative or less-informative by thresholding (experimentally kept $20$) the score $\rho$. Some sample scores are provided in Figure \ref{fig:Informativeness-of-captions}. The proposed strategy discards about $1.5 M$ ($55\%$) comments from the entire corpus. Subsequently, we remove the images which are left with no informative comments. Finally, we are left with $240,060$ images and $1,318,359$ comments, with an average of $5.58$ comments per image. We call this cleaner subset as \textbf{AVA-Captions} 
The proposed approach is evaluated by human subjects and the results are discussed in Figure~\ref{fig:sub_test_1} and Section \ref{subsec:subjective}. 
\section{Weakly Supervised CNN}
\label{subsec:Discovering-Aesthetic-Attributes}

Although the comments in AVA-Captions are cleaner than the raw comments, they cannot be directly used for training the CNN \ie the visual feature extractor. As discussed in Sec \ref{sec:introduction}, the standard approach followed in NIC and its variants is to use an ImageNet trained model for the task. In this section, we propose an alternative weakly supervised strategy for training the CNN from scratch by exploiting the \textit{latent} aesthetic information within the AVA-Captions. Our approach is motivated from two different areas: visual attribute learning and text document clustering.
\subsection{Visual and Aesthetic Attributes}
Visual Attribute Learning is an active and well-studied problem in computer vision. Instead of high-level object/scene annotations, models are trained for low-level attributes such as ``smiling face'', ``open mouth'', ``full sleeve'' \etc and the features are used for tasks such as image-ranking \cite{parikh2011relative}, pose-estimation \cite{zhang2014panda}, fashion retrieval \cite{wang2016walk}, zero-shot learning \cite{huang2015learning}, \etc. Similarly, our goal is to identify aesthetic attributes and train a CNN. A straightforward approach is to use the n-grams from comments (Sec \ref{subsec:Caption-filtering-strategy}) and use them as aesthetic attributes. However, there are two problems with this approach: Firstly, the set of n-grams is huge ($\sim25K$) and thus training the CNN directly using them as labels is not scalable. Secondly, several n-grams such as ``grayscale'', ``black and white'', ``bw'' refer to the same concept and carry redundant information. 

Therefore, we apply a clustering of semantically similar n-grams and thereby grouping the images which share similar n-grams in their corresponding comments. For example, portraits are more likely to contain attributes such as ``cute expression", ``face" \etc and landscape shots are more likely to share attributes such as ``tilted horizon", ``sky", ``overexposed clouds" \etc. Essentially, the intuition behind our approach is to discover clusters of photographic attributes or topics from the comment corpus and use them as labels for training the CNN. In text document analysis, it is a common practice to achieve such grouping of topics from a text corpus using a technique called Latent Dirichlet Allocation  \cite{blei2003latent}.

\begin{figure}
    \begin{center}
    \resizebox{.5\textwidth}{!}{
    \def\arraystretch{1.5}
        \begin{tabular}{p{0.2\textwidth}p{0.3\textwidth}}
        \hline
        \multicolumn{1}{c}{\textbf{Topics}} & \multicolumn{1}{c}{\textbf{Images}}\\
        \hline
        \hline
          ``Cute-Expression", ``Face", ``Ear"   & \multirow{5}{*}{\includegraphics[width = .3\textwidth]{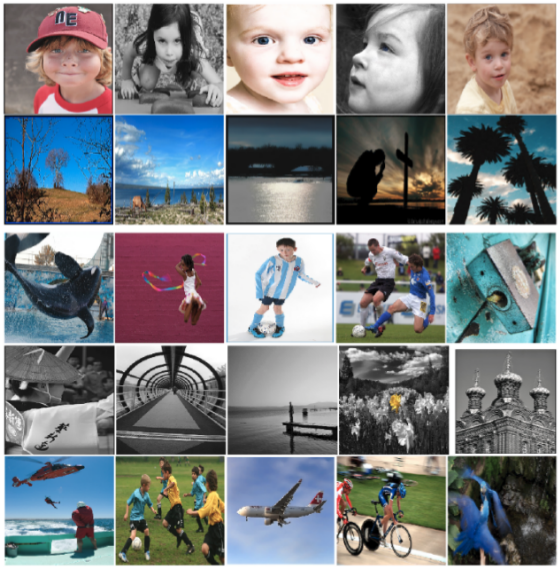}} \\
          \cline{1-1}
        ``Landscape",``Sky'', ``Cloud''     &\\
        \cline{1-1}
        ``Action Shot", ``Sport", ``Great Action" & \\
        \cline{1-1}
        ``Black and white", ``Tone", ``Contrast" & \\
        \cline{1-1}
        ``Motion Blur", ``Movement", ``Shutter Speed" & \\
        \hline
        \end{tabular}}
    \end{center}
    
    \caption{Some topics / labels discovered from AVA-Captions using LDA. }
    \label{fig:lda_topics}
\end{figure}
\subsection{Latent Dirichlet Allocation (LDA)}

LDA is an unsupervised generative
probabilistic model, widely used for topic modelling in text corpora.
It represents text documents as a probabilistic mixture of topics, and each topic as a probabilistic mixture of words. The words which co-occur frequently in the corpus are grouped together by LDA to form a topic. For example, by running LDA on a large corpus of news articles, it is possible to discover topics such as ``sports'', ``government policies'', ``terrorism'' etc \cite{lda-medium}. 

Formally stated, given a set of documents $D_{i}=\{D_{1},D_{2}...D_{N}\}$,
and a vocabulary of words $\omega_{i}=\{\omega_{1},\omega_{2}...\omega_{M}\}$,
the task is to infer K latent topics $T_{i}=\{T_{1,}T_{2,}\dots T_{K}\}$,
where each topic can be represented as a collection of words (term-topic
matrix) and each document can be represented as a collection of topics
(document-topic matrix). The term-topic matrix represents the probabilities of each word associated with a topic and the document-topic matrix refers to the distribution of a document over the $K$ latent topics. The inference is achieved using a variational Bayes approximation \cite{blei2003latent} or Gibb's sampling \cite{porteous2008fast}. A more detailed explanation can be found in \cite{blei2003latent}. 

\subsection{Relabelling AVA Images}
We regard all the comments corresponding to a given image as a document. The vocabulary is constructed by combining the unigrams and bigrams
extracted from the AVA-Captions as described in Section \ref{subsec:Caption-filtering-strategy}.
In our case: $N=230,698$ and $M=25,000$, and $K$ is experimentally
fixed to $200$. By running LDA with these parameters on AVA-Captions, we discover $200$ latent topics, composed of n-grams which co-occur
frequently. The method is based on the assumption that the visual aesthetic
attributes in the image are correlated with the corresponding comments
and images possessing similar aesthetic properties are described using
similar words. 

Even after the caption cleaning procedure, we observe that n-grams such as ``nice composition" or ``great shot" still occur more frequently than others. But, they occur mostly as independent clauses in bigger comments such as ``\textit{I like the way how the lines lead the eyes to the subject. Nice shot}!". In order to avoid inferring topics consisting of these less discriminative words, we consider only those n-grams in the vocabulary which occur in less than $10\%$ comments. 

In Figure \ref{fig:lda_topics}, we select $5$ topics thus inferred and some of the corresponding images whose captions belong to these topics. It can be observed that the images and the words corresponding to each topic are fairly consistent and suitable to be used as labels for training the CNN. 

\subsection{Training the CNN}
\begin{figure}
\begin{center}
\resizebox{.475\textwidth}{!}{
\def\arraystretch{1.75}
\begin{tabular}{|c|}
\hline 
\includegraphics[width=0.5\textwidth]{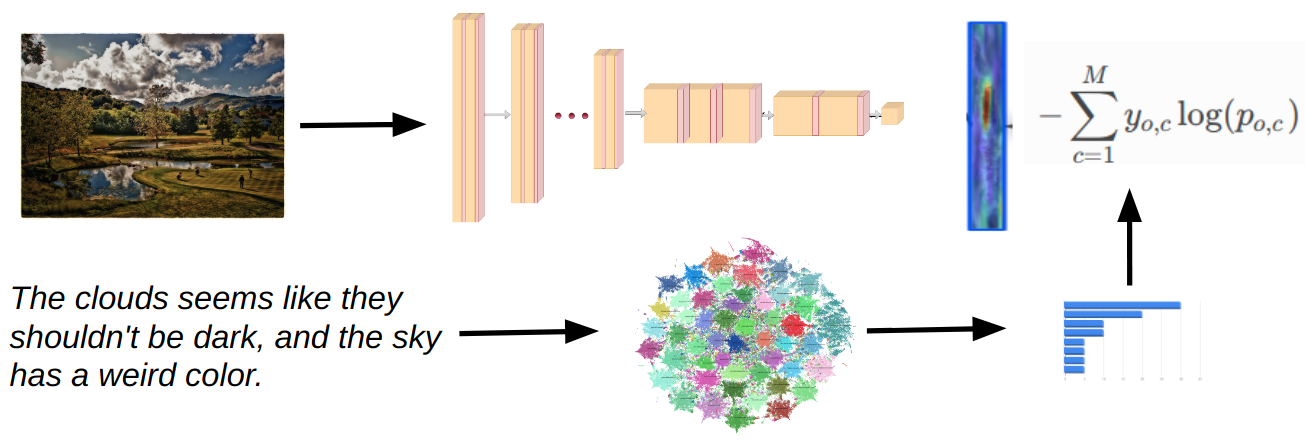}\\
\hline
\multicolumn{1}{p{0.5\textwidth}}{\textbf{ (a) Weakly-supervised training of the CNN: } Images and comments are provided as input. The image is fed to the CNN and the comment is fed to the inferred topic model. The topic model predicts a distribution over the topics which is used as a label for computing the loss for the CNN.}\\
\hline
 \includegraphics[width=0.5\textwidth]{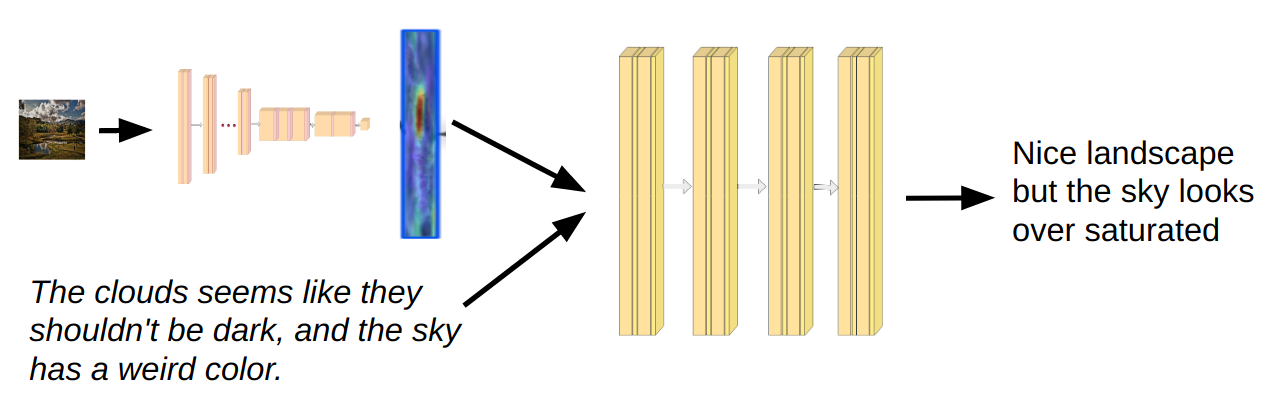} \tabularnewline
\hline 
\multicolumn{1}{p{0.5\textwidth}}{\textbf{ (b) Training the LSTM: } Visual features extracted using the CNN and the comment is fed as an input to the LSTM which predicts a candidate caption.}
\end{tabular}}
\end{center}
\caption{\textbf{Proposed pipeline}}
\label{fig:framework}
\end{figure}

Given an image and its corresponding captions, we estimate the
topic distribution $D_{T}$ of the comments. The CNN is trained
using $D_{T}$ as the ground-truth label. We adopt the ResNet101 \cite{he2016deep} architecture and replace the last fully connected layer with $K$ outputs, and train the framework using cross-entropy loss \cite{rubinstein2013cross} as shown in Figure \ref{fig:framework}a. 
\section{The Final Framework}
We adopt the NeuralTalk2 \cite{Luo2017} framework as our basis. Note, that our approach is generic and can be used with any CNN-LSTM framework for image captioning. In \cite{Luo2017}, visual features are extracted using an ImageNet trained ResNet101 \cite{he2016deep} which are passed as input to an LSTM for training the language model using the ground-truth captions. For our framework, we use two alternatives for visual features (a) ImageNet trained (b) weakly supervised (Sec \ref{subsec:Discovering-Aesthetic-Attributes}). The LSTM architecture is kept unchanged except hyper-parameters such as vocabulary size, maximum allowed length of a caption \etc. The language model is trained using the clean and informative comments from the AVA-Captions dataset (See Figure \ref{fig:framework}b).
\section{Experiments}
\label{sec:Experiments}
The experiments are designed to evaluate the two primary contributions: First, the caption
cleaning strategy and second, the weakly-supervised training of the
CNN. Specifically, we investigate: \textbf{(a)} the effect of caption filtering and the weakly supervised approach on the quality of captions generated in terms of accuracy (Sec \ref{subsec:accuracy}) and diversity (Sec \ref{subsec:diversity}), \textbf{ (b)} the generalizability of the captions learnt from AVA, when tested on other image-caption datasets (Sec \ref{subsec:generalizability}), \textbf{(c)} subjective or human opinion about the performance of the proposed framework (Sec \ref{subsec:subjective}).
\subsection{Datasets}
\textbf{AVA-Captions: } The compiled AVA-Captions dataset is discussed in detail in Section \ref{subsec:Caption-filtering-strategy}.  We use $230,698$ images and $1,318,359$ comments for training; and $9,362$ images for validation. 

\textbf{AVA raw-caption: } The original AVA dataset provided by Murray \etal \cite{murray2012ava} and the raw unfiltered comments are used to train the framework in order to observe the effects of caption filtering. 

\textbf{Photo Critique Captioning Dataset (PCCD): } This dataset was introduced by \cite{chang2017aesthetic}
and is based on www.gurushots.com. Professional photographers provide
comments for the uploaded photos on seven aspects: general impression,
composition and perspective, color and lighting, subject of photo,
depth of field, focus and use of camera, exposure and speed. In order to verify whether the proposed framework can generate aesthetic captions for images beyond the AVA dataset we trained it with AVA-Captions and tested it with PCCD.
For a fair comparison, we use the same validation
set provided in the original paper. 
\subsection{Baselines}
\label{sec:baselines}
We compare three implementations: \textbf{ (a) Noisy - Supervised (NS): } NeuralTalk2 \cite{Luo2017} framework trained on AVA-Original. It has an ImageNet trained CNN, followed by LSTM trained on raw, unfiltered AVA comments. NeuralTalk2 is also used as a baseline for AIC in \cite{chang2017aesthetic}. \textbf{(b) Clean - Supervised (CS): } The LSTM of the NeuralTalk2 is trained on AVA-Captions \ie filtered comments. The CNN is same as NS \ie Imagenet trained. \textbf{ (c) Clean and weakly-supervised (CWS): } NeuralTalk2 framework, where the CNN is trained with weak-supervision using LDA and the language model is trained on AVA-Captions. 
\subsection{Results and Analysis}
\label{sec:results}
\begin{table*}
\begin{center}
\def\arraystretch{1.2}
\resizebox{\textwidth}{!}{
\begin{tabular}{cc}
\resizebox{.6\textwidth}{!}{
\begin{tabular}{|p{1.5cm}|c|c|c|c|c|c|c|c|c|}
\hline 
\textbf{Method} & \textbf{B1} & \textbf{B2} & \textbf{B3} & \textbf{B4} & \textbf{M} & \textbf{R} & \textbf{C} & \textbf{S} & \textbf{S-1}\tabularnewline
\hline 
\hline 
NS & 0.379 & 0.219 & 0.122 & 0.061 & 0.079 & 0.233 & 0.038 & 0.044 & 0.135\tabularnewline
CS & 0.500  & 0.280  & 0.149 & 0.073 & 0.105 & 0.253 & \textbf{0.060} & \textbf{0.062} & \textbf{0.144} \tabularnewline
CWS & \textbf{0.535} & \textbf{0.282} & \textbf{0.150} & \textbf{0.074} & \textbf{0.107} & \textbf{0.254} & 0.059 & 0.061 & \textbf{0.144}\tabularnewline
\hline 
\end{tabular}
} &
\resizebox{.4\textwidth}{!}{
\begin{tabular}{|c|c|c|c|c|c|}
\hline 
\textbf{Method} & \textbf{Train} & \textbf{Val} &\textbf{S-1} & \textbf{P} & \textbf{R}\tabularnewline
\hline 
\hline 
CNN-LSTM-WD &PCCD&PCCD& 0.136 & 0.181 & 0.156\tabularnewline
AO&PCCD&PCCD&0.127 & 0.201 & 0.121\tabularnewline
AF &PCCD&PCCD& 0.150 & 0.212 & 0.157\tabularnewline
\hline 
CS &AVA-C&PCCD& 0.144 & 0.166 & 0.166\tabularnewline
CWS &AVA-C&PCCD& 0.142 & 0.162 & 0.161\tabularnewline
\hline 
\end{tabular}}\\
\textbf{(a) Accuracy} & \textbf{(b) Generalizability} \\
\end{tabular}
}
\end{center}
\caption{(a) \textbf{Results on AVA-Captions:} Both CS and CWS, trained on AVA-Captions perform significantly better than NS, which is trained on nosiy data. Also, the performance of CWS and CS is comparable, which proves the effectiveness of the weakly supervised approach (b) \textbf{Generalization results on PCCD:} Models trained on AVA-C perform well on PCCD validation set, when compared with models trained on PCCD directly. We argue that this impressive generalizability is achieved by training on a larger and diverse dataset.}
\label{tab:automated_metrics}
\end{table*}
\subsubsection{Accuracy}
\label{subsec:accuracy}
Most of the existing standards for evaluating
image captioning such as BLEU~(B) \cite{papineni2002bleu}, METEOR~(M) \cite{banerjee2005meteor},
ROGUE~(R) \cite{lin2004rouge}, CIDEr~(C) \cite{vedantam2015cider} etc. are
mainly more accurate extensions of the brute-force method \cite{cui2018learning} \ie comparing the n-gram overlap between candidate and reference captions.
Recently introduced
metric SPICE~(S) \cite{anderson2016spice} instead compares
scene graphs computed from the candidate and reference captions. It
has been shown that SPICE captures semantic similarity better and is closer to human judgement more than the
rest. Traditionally, SPICE is computed between the candidate and all the reference captions. A variant of SPICE (which we refer to as S-1) is used in \cite{chang2017aesthetic} where the authors compute SPICE between the candidate and each of the reference captions and choose the best. In this paper, we report both S and S-1.

From Table \ref{tab:automated_metrics}(a), we observe that both CS and CWS outperform NS significantly over all metrics. Clearly, training the framework with cleaner captions results in more accurate outputs. On the other hand, the performance of CWS and CS is comparable. We argue that this indicates that the proposed weakly-supervised training strategy is capable of training the CNN as efficiently as a purely supervised approach and extract meaningful aesthetic features. Additionally as mentioned in Sec \ref{sec:introduction}, the proposed CWS approach has an advantage that it does not require expensive human annotations to train. Thus, it is possible to scale to deeper architectures, and thus learn more complex representations simply by crawling the vast, freely-available and weakly-labelled data from the web. 
\subsubsection{Diversity}
\label{subsec:diversity}
\begin{figure*}
\centering
\resizebox{\textwidth}{!}{
\def\arraystretch{1.2}
\begin{tabular}{cccc}
\includegraphics[width=0.25\textwidth]{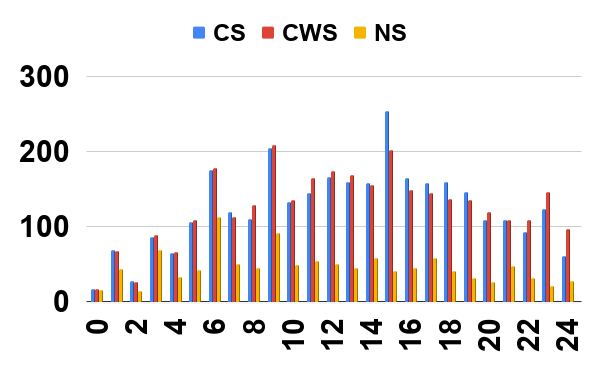} & \includegraphics[width=0.25\textwidth]{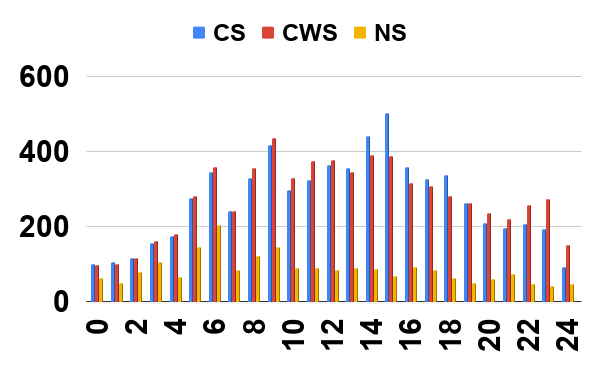} &
\includegraphics[width=0.25\textwidth]{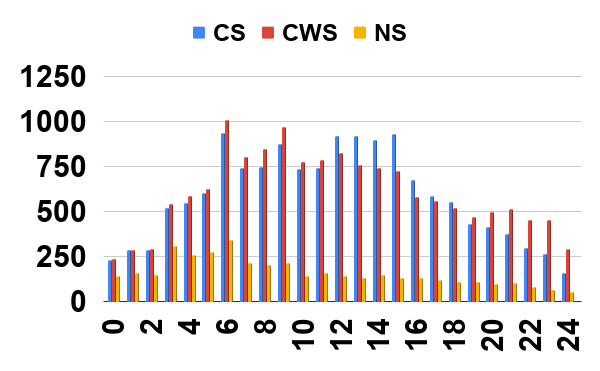} & \includegraphics[width=0.25\textwidth]{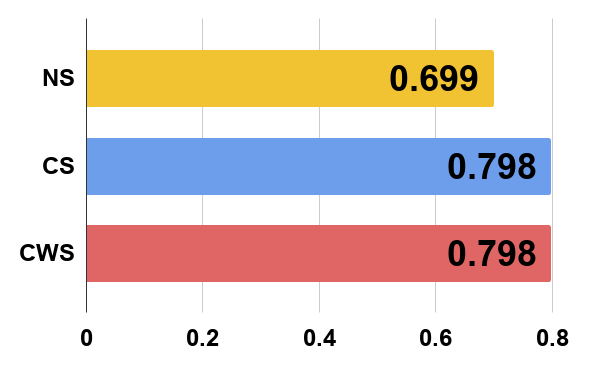}\\
(a) 1-gram & (b) 2-gram & (c) 4-gram & (d) Overall\tabularnewline
\end{tabular}
}
\caption{\textbf{Diversity:} Figures (a) - (c) report diversity of captions following \cite{aneja2018convolutional}. The $x$-axes correspond to n-gram positions in a sentence. The $y$-axes correspond to the number of unique n-grams at each position, for the entire validation set. Figure (d) plots the overall diversity, as reported in \cite{chang2017aesthetic}.  We observe that the diversity of the captions increase significantly when the framework is trained on cleaner ground-truth \ie AVA-Captions (CS or CWS) instead of AVA-Original (NS).}
\label{fig:diversity}
\end{figure*}
Image Captioning pipelines often suffer from monotonicity of captions \ie similar captions are generated for the validation images.
This is attributed to the fact that the commonly used cross-entropy
loss function trains the LSTM by reducing the entropy of the output
word distribution and thus giving \emph{a peaky} posterior probability
distribution \cite{aneja2018convolutional}. As mentioned earlier in Section \ref{sec:introduction}, this is more pronounced in AIC due to the vast presence of the \textit{easy} comments in the web. Diversity of the captions is usually measured by overlap between the candidate and the reference captions. We evaluate diversity following two state-of-the-art
approaches~\cite{chang2017aesthetic,aneja2018convolutional}. In \cite{chang2017aesthetic}, the authors define that
two captions are different if the ratio of common words between them
is smaller than a threshold ($3\%$ used in the paper).
In \cite{aneja2018convolutional},
from the set of all the candidate captions, the authors compute the
number of unique n-grams ($1,2,4$) at each position starting from
the beginning up to position 13.

We plot diversity using \cite{chang2017aesthetic} in Figure \ref{fig:diversity}d.  We compute using the alternative approach of \cite{aneja2018convolutional} in Figure \ref{fig:diversity}(a-c) but up to 25 positions since on an average the AVA captions are longer than the COCO captions. From both, we notice that diversity of NS is significantly lesser than CS or CWS. We observe that NS ends up generating a large number of ``safe" captions such as ``I like the composition and colours" or ``nice shot" \etc. We argue, that our caption filtering strategy reduces the number of useless captions from the data and thus the network learns more accurate and informative components.
\subsubsection{Generalizability}
\label{subsec:generalizability}
We wanted to test whether the knowledge gained by training on a large-scale but weakly annotated dataset is generic \ie transferable to other image distributions. To do so,  we train our frameworks on AVA-Captions and compare them with the models from \cite{chang2017aesthetic}, trained on PCCD.  Everything is tested on the PCCD validation set. 
The models used by \cite{chang2017aesthetic} are: (a) CNN-LSTM-WD is the NeuralTalk2 framework trained on PCCD. (b) Aspect oriented (AO) and (c) Aspect fusion (AF) are supervised methods, trained on PCCD. Note, that all the models are based on the NeuralTalk2 framework \cite{Luo2017} and hence comparable in terms of architecture.  

In Table \ref{tab:automated_metrics}(b), we observe that both CS and CWS outperform CNN-LSTM-WD and AO in S-1 scores. AF is still the best strategy for the PCCD dataset. 
Please note, both AO and AF are supervised strategies and require well defined ``aspects" for training the network. Hence, as also pointed out in \cite{chang2017aesthetic}, it is not possible to train these frameworks on AVA as such aspect-level annotations are unavailable. However, we observe that both CS and proposed CWS, trained on AVA-Captions score reasonably well on PCCD. They are also generic strategies which can be easily mapped to other captioning tasks with weak supervision. We argue that the observed generalization capacity is due to training with a large and diverse dataset.
\subsubsection{Subjective (Human) Evaluation}
\label{subsec:subjective}
\begin{figure}
\begin{center}
\resizebox{.5\textwidth}{!}{
\def\arraystretch{1.2}
\begin{tabular}{cc}
\includegraphics[width=0.25\textwidth]{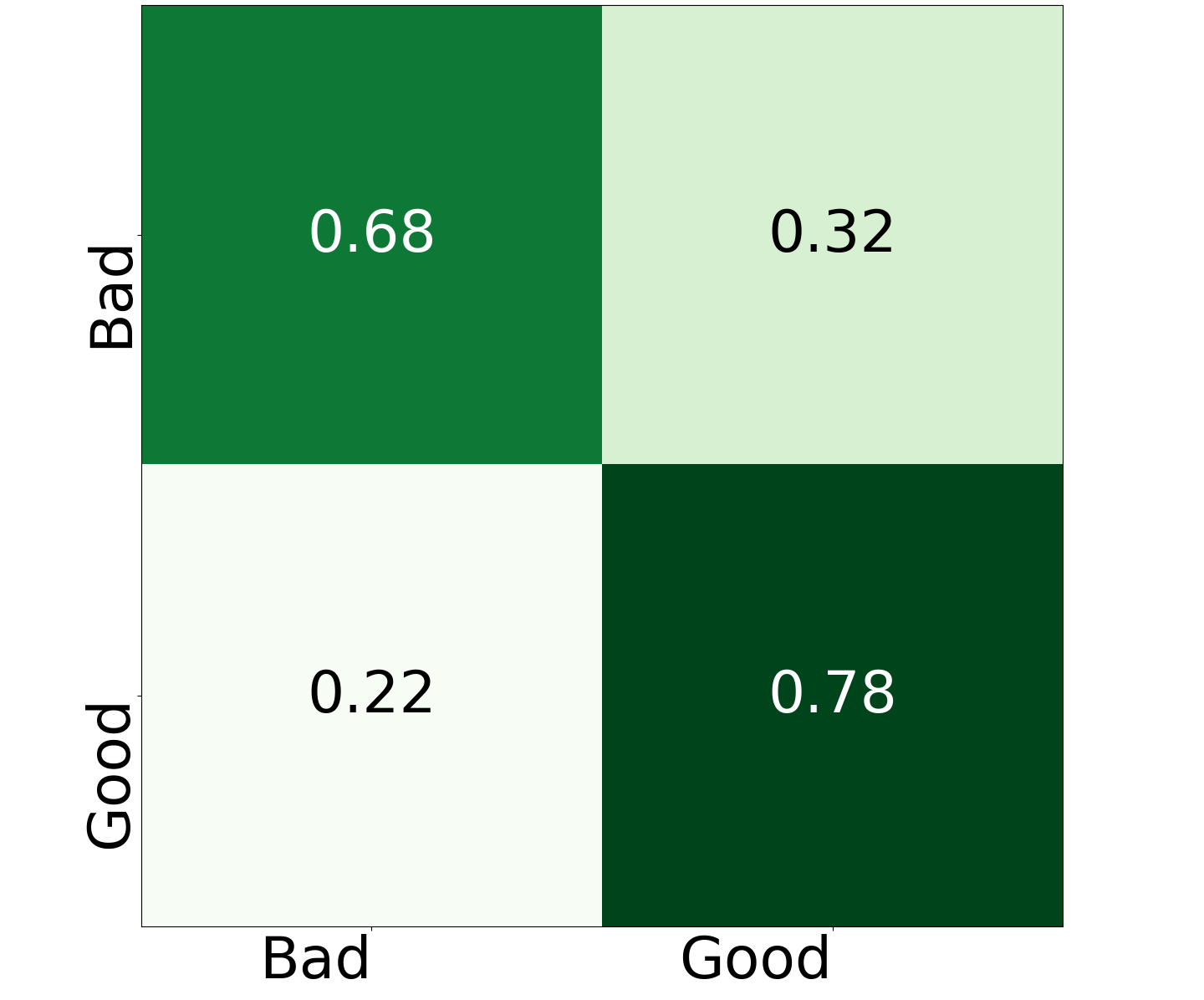} & \includegraphics[width=0.25\textwidth]{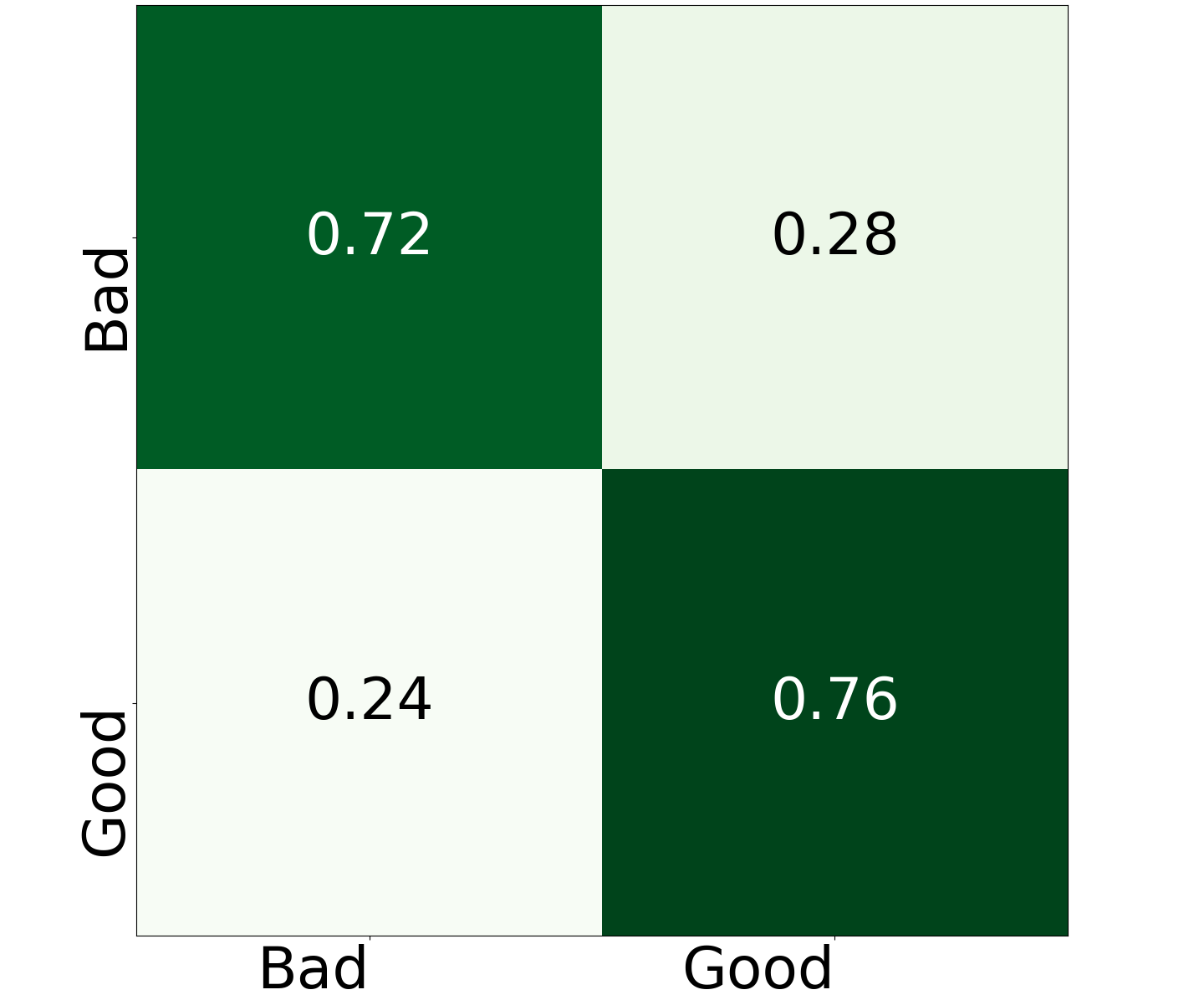}\tabularnewline
(a) Experts & (b) Non Experts\tabularnewline
\end{tabular}}
\end{center}
\caption{\textbf{Subjective evaluation of caption filtering: } The matrix compares our scoring strategy and human judgement for distinguishing a \textit{good} and a \textit{bad} caption. The rows stand for our output, and the columns represent what humans thought. We observe that the proposed caption filtering strategy is fairly consistent with what humans think about the informativeness of a caption.}
\label{fig:sub_test_1}
\end{figure}
Human judgement is still the touchstone for evaluating image captioning, and all the previously mentioned metrics are evaluated based on how well they correlate with the same. Therefore, we perform quality assessment of the generated captions by a subjective study. Our experimental procedure
is similar to Chang \etal \cite{chang2017aesthetic}. We found $15$ participants with varying degree of expertise in photography
($4$ experts and $11$ non-experts) to evaluate our framework. In order to familiarize the participants with the grading process, a brief training with $20$ trials was provided
beforehand. The subjective evaluation was intended to assess: (a) whether the caption scoring strategy (Equation \ref{eq:informativeness}) is consistent with human judgement regarding the same (b) the effect of cleaning on the quality of generated captions.

\textbf{(a) Consistency of Scoring Strategy:} 
We chose $25$ random images from
the validation set, and from each image, we selected $2$ accepted and $2$ discarded captions. During the experiment the subject was shown an image and a caption, and was asked to give a score on a scale of $100$. In Figure \ref{fig:sub_test_1}a and \ref{fig:sub_test_1}b, we plot our predictions and human judgement in a confusion matrix. We find that our strategy is fairly consistent with what humans think as a good or a bad caption. Interestingly, with the experts, our strategy produces more false positives for bad captions. This is probably due to the fact that our strategy scores long captions higher, which may not always be the case and is a limitation. 

\textbf{(b) Effect of Caption Filtering: } 
Similarly, $25$ random images were chosen
from the validation set. Each image had 3 captions, the candidates generated by NS, CS and CWS frameworks. During each
trial, the subject was shown an image and one of the captions and
asked to rate it into one of the categories - Good, Average and Bad.
These categories follow from \cite{chang2017aesthetic} and the participants were asked to rate a caption based on whether it conveyed enough information about a photograph. We observe in Table \ref{tab:sub_2} the percentage of good, common and bad captions generated by each method.

We observe that humans did not find any caption from NS to be good. Most of them were common or bad. This is due to its high tendency to generate the short, safe and common captions. Humans find CS to be performing slightly better than CWS which can probably be attributed to the lack of supervision during training the CNN. But as mentioned in Section~\ref{sec:introduction}, semi-supervised training is effective in practical scenarios due the easy availability of data and it might be worth investigating whether it is possible to improve its performance using more data and more complex representations. Additional qualitative results are provided in Figure \ref{fig:title_fig} and also the supplementary material.
\begin{table}
\begin{center}
\resizebox{.475\textwidth}{!}{
\def\arraystretch{1.2}
\begin{tabular}{|c|p{.5cm}p{.5cm}p{.5cm}|c|p{.5cm}p{.5cm}p{.5cm}|c|}
\cline{2-9} 
\multicolumn{1}{c}{} & \multicolumn{4}{|c|}{\textbf{Experts}} & \multicolumn{4}{c|}{\textbf{Non-Experts}}\tabularnewline
\hline 
 \textbf{Method} & \textbf{Good} (3) & \textbf{Com} (2) & \textbf{Bad} (1) & \textbf{Avg} & \textbf{Good} (3) & \textbf{Com} (2) & \textbf{Bad} (1) & \textbf{Avg}\tabularnewline
\hline 

NS & 0 & 80 & 20 & 1.80 & 0 & 84 & 16 & 1.84\tabularnewline

CS & 8 & 84 & 8 & 2.0 & 28 & 68 & 4 & 2.24\tabularnewline
CWS & 4 & 80 & 16 & 1.88 & 20 & 72 & 8 & 2.12\tabularnewline
\hline 
\end{tabular}
}
\end{center}
\caption{\textbf{Subjective comparison of baselines: } We observe that human subjects find CS and CWS to be comparable but both significantly better than NS. This underpins the hypothesis derived from the quantitative results that filtering improves the quality of generated captions and the weakly supervised features are comparable with the ImageNet trained features}
\label{tab:sub_2}
\end{table}
\section{Conclusion}
In this work, we studied aesthetic image captioning which is a
variant of natural image captioning. The task is challenging not only due to its inherent subjective nature but also due to the absence of a suitable dataset. To this end, we propose a strategy to clean the weakly annotated data easily available from the web and compile AVA-Captions, the first large-scale dataset for aesthetic image captioning. Also, we propose a new weakly-supervised approach to train the CNN. 
We validated the proposed framework thoroughly, using automatic metrics and subjective studies. 

As future work, it could be interesting to explore alternatives for utilizing the weak-labels and exploring other weakly-supervised strategies for extracting rich aesthetic attributes from AVA. It could also be interesting to extend this generic approach to other forms of captioning such as visual storytelling \cite{Kiros2015} or stylized captioning \cite{mathews2018semstyle} by utilizing the easily available and weakly labelled data from the web.\footnote{This publication has emanated from research conducted with the financial support of Science Foundation Ireland (SFI) under the Grant Number 15/RP/2776}
{\small
\bibliographystyle{ieee_fullname}
\bibliography{ms}
}

\end{document}